\documentclass[11pt]{article}

\usepackage[preprint]{acl}

\usepackage{times}
\usepackage{latexsym}
\usepackage{amsmath, amssymb}
\usepackage{graphicx}
\usepackage{enumitem}
\usepackage{booktabs}
\usepackage{colortbl}
\usepackage{xcolor}
\definecolor{winlow}{RGB}{255,220,220}
\definecolor{winmid}{RGB}{255,255,200}
\definecolor{winhigh}{RGB}{210,255,210}
\usepackage{subcaption}

\newcommand{\wincolor}[1]{%
    \ifdim #1 pt > 68pt
        \cellcolor{winhigh}#1%
    \else\ifdim #1 pt > 65pt
        \cellcolor{winmid}#1%
    \else
        \cellcolor{winlow}#1%
    \fi\fi
}

\usepackage[T1]{fontenc}

\usepackage[utf8]{inputenc}

\usepackage{microtype}

\usepackage{inconsolata}

\usepackage{graphicx}

\title{MixDPO: Modeling Preference Strength for Pluralistic Alignment}

\author{
Saki Imai$^{1,*}$ \quad
Pedram Heydari$^{2,*}$ \quad
Anthony Sicilia$^{3}$ \\
{\bf Asteria Kaeberlein}$^{1}$ \quad
{\bf Katherine Atwell}$^{1}$ \quad
{\bf Malihe Alikhani}$^{1}$ \\
\\
$^{1}$Northeastern University \quad
$^{2}$Johns Hopkins University \quad
$^{3}$West Virginia University \\
\\
$^{*}$Equal contribution
}

\begin{document}
\maketitle
\begin{abstract}
Preference based alignment objectives implicitly assume that all human preferences are expressed with equal strength. In practice, however, preference strength varies across individuals and contexts—a phenomenon established in behavioral economics and discrete choice theory. This mismatch limits the ability of existing objectives to faithfully capture heterogeneous human judgments. Inspired by this literature, we introduce \textbf{Mix}ed Logit \textbf{D}irect \textbf{P}reference \textbf{O}ptimization (MixDPO), a generalization of Direct Preference Optimization that models variation in preference strength. MixDPO enables alignment objectives to capture heterogeneity in how strongly preferences are expressed across training examples.
We evaluate MixDPO on three preference datasets using two open-weight language models. Across datasets, MixDPO improves aggregate alignment performance (+11.2 points on Pythia-2.8B) while preserving subgroup level preferences, with the largest gains appearing in settings with higher inferred preference heterogeneity. MixDPO makes preference heterogeneity explicit through learned strength distributions. We release our code for reproducibility.
\end{abstract}

\begin{figure}[!ht]
    \centering
    \includegraphics[width=0.99\linewidth]{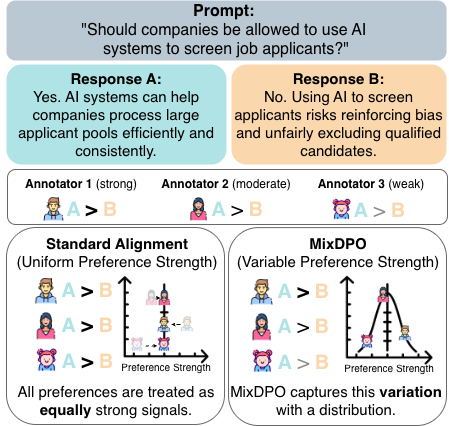}
    \caption{\textbf{Motivating example of latent preference strength heterogeneity.} Although annotators provide only binary comparisons, variation in repeated preference judgments—holding preference direction fixed—can arise from latent differences in how sharply individuals distinguish between alternatives. Mixed logit models capture this variation through a random sensitivity parameter.}
    \label{fig:fig1}
\end{figure}

\section{Introduction}
Preference based alignment methods \cite{rafailov2023direct, ouyang2022training} train language models from human comparisons, implicitly assuming that all preferences exert equal influence during training. However, human preferences are inherently heterogeneous. Fig.~\ref{fig:fig1} illustrates a scenario in preference based alignment. Given a prompt \textit{``Should companies be allowed to use AI to screen job applicants?''}, annotators compare two responses emphasizing different considerations such as efficiency (A) versus fairness (B). Although annotators may agree on which response they prefer, they often differ in how strongly that preference is expressed. Importantly, preference strength is not directly observed in alignment datasets. Instead, we model it as a latent random variable whose distribution is inferred from variation in pairwise comparisons, following the mixed logit tradition.

We argue that alignment under heterogeneous preferences requires modeling preference strength itself as a distribution. The existence of heterogeneous preference strength is a finding in behavioral economics and decision theory that long predates AI alignment \cite{mcfadden1972conditional, domencich1975urban}. Mixed logit models were developed to capture this phenomenon by modeling preferences as draws from a distribution rather than assuming a fixed sensitivity \cite{mcfadden2000mixed}. 

Pluralistic datasets used for language model alignment inherit this heterogeneity through controversial or value guided prompts \cite{kirk2024prism, zhang2025cultivating}. Despite this, existing preference based alignment objectives assume a global preference strength, effectively treating all preference comparisons as equally important. A fixed sensitivity parameter couples frequency with influence: preference comparisons that occur more frequently dominate the training gradient. As a result, frequently occurring but weakly expressed preferences can outweigh infrequent but strongly expressed ones, causing aggregate improvements to mask regressions on underrepresented groups.

We introduce \textbf{Mix}ed Logit \textbf{D}irect \textbf{P}reference \textbf{O}ptimization (MixDPO), which relaxes this assumption by modeling preference strength as a variable drawn from a learned distribution. MixDPO automatically downweights ambiguous or weakly expressed preferences while preserving strong signals, without requiring subgroup labels or heuristic reweighting. This formulation makes preference heterogeneity explicit and measurable via learned distributional parameters.

We evaluate MixDPO on three preference datasets (Anthropic HH, PRISM, and Community Alignment) and investigate two questions. First, we examine whether \textit{aggregate evaluation metrics mask subgroup level alignment differences}, by downweighting underrepresented subpopulations. Second, we assess whether \textit{MixDPO objectives preserve subgroup preferences without sacrificing aggregate performance}. On PRISM, which is explicitly designed to capture value diversity, MixDPO achieves higher subgroup averaged preference margins across dimensions such as age, gender, education, and employment status, and conversation type, while remaining competitive under aggregate metrics.


\paragraph{Contributions.} We show that \textit{aggregate alignment metrics can mask subgroup level differences} by disproportionately weighting frequent preference comparisons (\S~\ref{ssec:aggregated}). We introduce \textbf{MixDPO}, which allows preference comparisons to exert varying strengths of influence during optimization by modeling preference strength as a distribution(\S~\ref{sec:mldpo}). We demonstrate that this \textit{distributional modeling} leads to better preservation of subgroup preferences while maintaining competitive aggregate alignment performance (\S~\ref{ssec:h2}). Finally, we show that MixDPO captures dataset level preference heterogeneity and naturally collapses to fixed strength behavior when preferences are homogeneous (\S~\ref{ssec:distributional}).
\section{Background}
\label{sec:background}
\subsection{Preference Optimization as Probabilistic Choice}
Preference based alignment methods can be viewed as learning a probabilistic choice model over pairs of model outputs. Given a prompt $x$ and two candidate responses $(y_w, y_l)$, a preference annotation indicates that $y_w$ is preferred to $y_l$. Methods such as DPO \cite{rafailov2023direct}, KTO \cite{ethayarajh2024kto}, IPO \cite{azar2024general} and SimPO \cite{meng2024simpo} model this preference as arising from a reward function, with the probability of preferring $y_w$ over $y_l$ increasing with the reward difference.

Under a standard logistic choice model, this probability takes the form
$P(y_w \succ y_l \mid x) = \sigma\big(\beta \, \Delta r\big)$ where $\Delta r$ is the reward difference between the two responses and $\beta > 0$ controls the sensitivity of the choice probability to this difference. Larger values of $\beta$ yield more deterministic preferences while smaller values correspond to weaker preferences. In this sense, $\beta$ controls the sensitivity of observed preferences to differences in reward. In existing alignment methods, $\beta$ is typically treated as a fixed scalar shared across all training examples. This implies that all preference annotations are assumed to reflect the same level of confidence, consistency, or decisiveness, regardless of prompt, annotator, or context.

\subsection{Preference Heterogeneity and Pluralism}
Empirically, human preferences are not expressed with uniform strength \cite{mcfadden1972conditional}. Preferences are known to be heterogeneous, context dependent, and often stochastic rather than fixed \cite{slovic1995construction}. From the perspective of social choice theory, aggregating diverse and conflicting preferences has been recognized as a challenge, with no single aggregation rule capable of fully representing pluralistic values \cite{Arrow1951-ARRIVA}.

Recent work in machine learning has framed these concerns under the notion of \emph{pluralistic alignment}, which emphasizes the need for systems to reflect diverse human values rather than optimizing for a single averaged objective \cite{sorensen2024roadmap}. Existing approaches have explored preference inputs that augment binary comparisons with multi-attribute annotations \cite{li2024aligning}, dispersion based modeling \cite{chen2025mallowspo}, and ideal preference points \cite{chen2025pal}. While these methods address diversity in preference direction, they generally assume that preferences are expressed with uniform strength or reliability.

\subsection{Mixed Logit Models for Preference Heterogeneity}
Mixed logit models (also known as the random-coefficients logit) represent heterogeneity in choice behavior by assuming that each decision instance is generated by a sample from a distribution over preference strength \cite{mcfadden1972conditional}. Instead of assuming a fixed sensitivity coefficient, mixed logit treats such coefficients as random variables drawn from a population level distribution. \citet{mcfadden2000mixed} show that mixed logit can approximate any random utility model arbitrarily well under mild regularity conditions. We provide additional background on mixed logit models and their use for modeling heterogeneous preferences in Appendix~\ref{app:mixed_logit}.

In the context of preference optimization, this corresponds to treating the preference sensitivity $\beta$ as a random variable drawn from a distribution $p(\beta)$. The mean of this distribution captures the typical strength of preferences in the dataset, while its variance reflects the degree of heterogeneity in how strongly preferences are expressed. Standard fixed $\beta$ models are recovered as a special case when the distribution collapses to a point mass. Our work draws on this framework to model a distribution of preference strength in the alignment objective.

\subsection{Other Related Works}
Several recent works have explored relaxing the fixed $\beta$ assumption within the preference optimization framework. $\beta$-DPO adapts the sensitivity parameter $\beta$ at the batch level based on data quality \cite{wu2024beta}. \citet{hong2024adaptive} propose an adaptive preference scaling method adjusts the sensitivity of preference pairs to reflect uncertainty or ambiguity in the observed comparisons. While these approaches introduce adaptivity into preference optimization, sensitivity is treated as a deterministic quantity rather than as a random variable governed by a population level distribution.

While Proximal Policy Optimization (PPO) \cite{schulman2017proximal} dynamically regulate update magnitudes during training through an adaptive KL penalty coefficient, this regulation occurs at the level of optimization rather than preference modeling. Thus, it does not model variation in preference strength within the choice process itself.

\section{Mixed Logit Direct Preference Optimization}
\label{sec:mldpo}

Section~\ref{sec:background} motivates treating preference sensitivity as a source of heterogeneity rather than a fixed constant. We introduce MixDPO, which models preference strength as a random variable, allowing the alignment objective to capture heterogeneity in preference strength.

\subsection{Standard DPO}
Given a prompt $x$ and a preference pair $(y_w, y_l)$, Direct Preference Optimization (DPO) optimizes the objective
\begin{equation}\small
\mathcal{L}_{\text{DPO}}(\theta)
= -\mathbb{E}_{(x,y_w,y_l)}\left[
\log \sigma\big(\beta\,\Delta r_\theta\big)
\right],
\end{equation}
where $\Delta r_\theta = r_\theta(x,y_w) - r_\theta(x,y_l)$, and the implicit reward is defined as
\begin{equation}\small
r_\theta(x,y) = \log \frac{\pi_\theta(y \mid x)}{\pi_{\text{ref}}(y \mid x)}.
\end{equation}
The scalar $\beta > 0$ controls the strength of the preference signal and is fixed across all training examples.

\subsection{Mixed Logit DPO Objective}
MixDPO generalizes DPO by treating the preference strength $\beta$ as a random variable drawn from a distribution $p(\beta)$. The resulting objective marginalizes over preference strength:
\begin{equation}\small
\mathcal{L}_{\text{MixDPO}}(\theta)
= -\mathbb{E}_{(x,y_w,y_l)}\left[
\log \mathbb{E}_{\beta \sim p(\beta)}
\sigma\big(\beta\,\Delta r_\theta\big)
\right].
\end{equation}
This formulation allows the alignment objective to represent variation in how strongly preferences are expressed across examples, without conditioning on explicit subgroup or demographic labels.

\subsection{Learning the Preference Strength Distribution}
We parameterize the preference strength distribution $p(\beta)$ with parameters $\theta_\beta$, which are optimized jointly with the policy parameters $\theta$. We consider two families for $p(\beta)$. 

\paragraph{LogNormal.}
The LogNormal distribution is a standard choice in mixed logit models when coefficients must be strictly positive \cite{train2009, Hensher_Rose_Greene_2005}. We parameterize $\beta$ using location and scale parameters $(\mu, \sigma)$ and optimize them jointly with the policy. We use a reparameterized form $\beta = \exp(\mu + \sigma \varepsilon), \varepsilon \sim \mathcal{N}(0,1)$ which allows gradients to propagate through sampled preference strengths through backpropagation \cite{Kingma14,pmlr-v32-rezende14}. Since the resulting expectation $\mathbb{E}_{\beta}[\sigma(\beta \Delta r_\theta)]$ does not admit a closed form, it is approximated using Monte Carlo sampling (with $K{=}16$ samples per update). While this approach yields unbiased gradient estimates, it introduces gradient noise due to sampling. Full details of the Monte Carlo approximation and reparameterization are provided in Appendix~\ref{app:mc_reparam}.

\paragraph{Gamma.}
In contrast, when $\beta \sim \Gamma(k, \lambda)$, the inner expectation in the MixDPO objective admits a closed form analytic solution. This eliminates the need for Monte Carlo sampling and enables deterministic gradient propagation through both the policy parameters $\theta$ and the distribution parameters $\theta_\beta$. Formally, for $\beta_0 \sim \Gamma(k, \lambda)$ with density $f(\beta)=\frac{\lambda^k}{\Gamma(k)}\beta^{k-1}e^{-\lambda\beta}$, the inner expectation admits a closed form analytical solution (see Appendix~\ref{app:closedform_derivation} for derivation). When $\Delta r_\theta>0$,
\begin{equation}\small
\mathbb{E}_{\beta_0 \sim p(\beta_0)}
=
1 - \Big(\frac{\lambda}{\Delta r_\theta}\Big)^{k}
\Phi\!\Big(-1,k,1+\frac{\lambda}{\Delta r_\theta}\Big)
\end{equation}
and when $\Delta r_\theta<0$,
\begin{equation}\small
\mathbb{E}_{\beta_0 \sim p(\beta_0)} =
\Big(\frac{\lambda}{-\Delta r_\theta}\Big)^{k}
\Phi\!\Big(-1,k,1+\frac{\lambda}{-\Delta r_\theta}\Big)
\end{equation}
where $\Phi(\cdot)$ denotes the Lerch transcendent function. For $\Delta r_\theta=0$, $\mathbb{E}_{\beta_0 \sim p(\beta_0)}=\frac{1}{2}$.

The Lerch transcendent $\Phi(z,s,a)$ admits a convergent series representation when
$|z|\le 1$ and $\Re(s)>1$. In our setting, these conditions are automatically
satisfied. Specifically, $z$ is fixed to $-1$, and the exponent $s$ corresponds to the Gamma shape parameter $k$, which is strictly positive. When $z=-1$, the Lerch transcendent reduces to a difference of Hurwitz zeta functions:
\begin{equation}\small
  \Phi(-1,s,a) = 2^{-s}\Big[\zeta(s,\frac{a}{2}) - \zeta(s,\frac{a+1}{2})\Big]
  \label{eq:lerch-hurwitz}
\end{equation}
where $\zeta(s, a)=\sum_{n=1}^\infty \frac{1}{(a+n)^s}$ denotes the Hurwitz zeta 
function. In practice, evaluating the resulting expectation therefore reduces to computing Hurwitz zeta functions, for which standard automatic differentiation frameworks do not provide gradient support. We therefore approximate numerically using a truncated series expansion ($n=1000)$ chosen to achive high numerical accuracy while preserving differentiability. Implementation details and error analysis are provided in Appendix~\ref{app:gamma_numerics}.
\section{Experiments}
This section describes the experimental setup used to compare DPO and MixDPO. We detail the datasets, models, and evaluation metrics.

\subsection{Datasets}
\label{ssec:datasets}
We study preference optimization under datasets that differ in the degree of
\emph{observable preference heterogeneity}, arising from both annotator diversity
and the diversity of values elicited by the prompts.

\paragraph{PRISM} is designed to capture value diversity in human judgments, containing over 1{,}500 annotators with associated demographic metadata and value profiles \cite{kirk2024prism}. The dataset includes multi-turn preference judgments across different conversational framings, including \textit{Unguided}, \textit{Values-Guided}, and \textit{Controversy} prompts. We treat conversational framing as a form of contextual subgroup and include it alongside demographic attributes (age, gender, education, employment status) in our disaggregated analysis.

\paragraph{Community Alignment} focuses on cross-cultural and community specific alignment, with preference judgments spanning multiple languages and cultural contexts \cite{zhang2025cultivating}. Prompts are designed to elicit value-laden judgments along established cultural value dimensions \cite{hofstede2011dimensionalizing, inglehart2005modernization}. In this work, we restrict our analysis to English examples and use the available metadata for subgroup analysis.

\paragraph{Anthropic HH} is a widely used preference dataset constructed to evaluate helpfulness and harmlessness in language model responses \cite{bai2022training}. The dataset does not include explicit subgroup or demographic annotations. We therefore treat HH as a \emph{relatively preference homogeneous control case} along these axes.

Additional details on annotator pool size and prompt diversity across datasets, and their implications for preference heterogeneity, are provided in Appendix~\ref{app:dataset_heterogeneity}.

\subsection{Models}
We conduct experiments using two open weight language models: Pythia-2.8B and Llama3.2-1B. All alignment methods use identical batch sizes, optimizer settings, learning rates for policy parameters, and maximum sequence lengths. Full training hyperparameters, optimization settings, and implementation details
are provided in Appendix~\ref{app:training_details}.

\subsection{Preference Strength Parameterization}
For DPO, we use $\beta=0.1$, which has been shown to perform well across prior work and is commonly used as a default setting.
For MixDPO, we experiment with both LogNormal and Gamma distributions over the preference strength parameter $\beta$. To assess the sensitivity of learning preference heterogeneity to optimization choices, we train the distributional parameters $\theta_\beta$ using learning rates $\{10^{-3}, 10^{-4}, 10^{-5}\}$. In addition, we include a \emph{fixed parameters} baseline in which $\beta$ is sampled from a distribution whose parameters are not optimized. This baseline isolates the effect of \emph{learning} the preference strength distribution. Specifically, the fixed distributions are parameterized as $\beta \sim \text{LogNormal}(\mu = -2.3, \sigma = 0.6)$ and $\beta \sim \text{Gamma}(k = 2.0, \lambda = 16.7)$ both yielding $\mathbb{E}[\beta] \approx 0.1$, matching the fixed $\beta$ used in standard DPO. For learned MixDPO variants, the distributional parameters are initialized from these fixed settings.

\begin{figure*}
    \centering
    \includegraphics[width=0.9\linewidth]{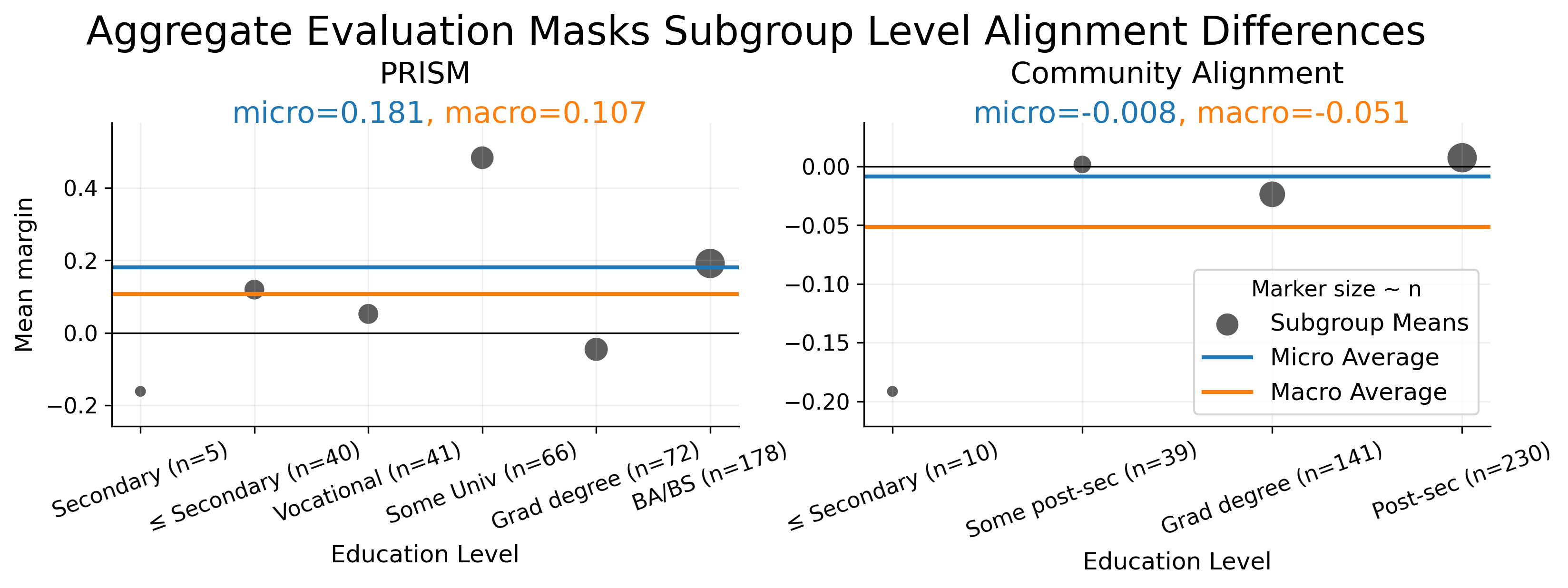}
    \caption{\textbf{Education subgroup mean preference margins for PRISM and Community Alignment.} Micro averages weight subgroups by frequency and can mask degraded alignment for underrepresented groups; macro averages assign equal weight per subgroup and better reflect subgroup conditioned performance. Marker size indicates subgroup sample size.}
    \label{fig:education_micro_macro}
\end{figure*}


\subsection{Evaluation Metrics}
For each dataset, we randomly sample 425 examples from the held out test set and evaluate alignment performance using the following metrics.

\paragraph{Win Rate}
We evaluate overall alignment performance using AlpacaEval 2.0 win rate and length-controlled win rate (LC-WR) \cite{dubois2024length}. For each test prompt, models generate responses that are compared against a baseline using GPT-4-Turbo as an evaluator, which selects the preferred completion. As a baseline, we use the supervised fine-tuned (SFT) model trained on the same dataset, to allow us to isolate the effect of preference based alignment on generation quality. Prior work has shown that AlpacaEval win rates exhibit high correlation with human judgments \cite{dubois2024length}.

\paragraph{Preference Margin}
While win rate provides a high level measure of generation quality, it depends on decoding and LLM evaluation, which can introduce variability. To obtain a deterministic assessment of alignment, we compute the log probability difference between preferred and dispreferred responses for each human labeled preference pair \cite{meng2024simpo}. This \emph{preference margin} is computed directly from the policy model. As such, it provides a human grounded measure of whether the policy has internalized the preference signals present in the training data.


\begin{figure*}[t]
    \centering
    \includegraphics[width=0.85\linewidth]{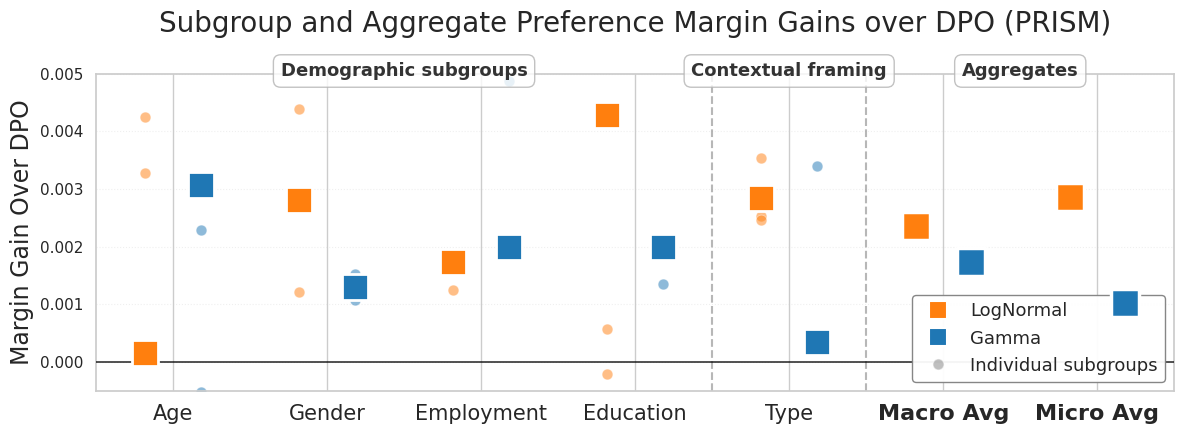}
    \caption{\textbf{Preference margin gains over DPO on PRISM (Pythia-2.8B)}. The figure reports subgroup level preference margin gains for demographic subgroups (annotator heterogeneity) and conversational framing (contextual heterogeneity), along with micro and macro averaged margins.
    MixDPO (LogNormal and Gamma) improve subgroup averaged (macro) preference margins relative to standard DPO while maintaining comparable micro averaged performance. This suggests improved preservation of subgroup preferences without aggregate tradeoffs.}
    \label{fig:prism_subgroup_margins}
\end{figure*}

\begin{table*}[t]
\centering
\setlength{\tabcolsep}{3.5pt}
\small
\resizebox{\textwidth}{!}{
\begin{tabular}{llcccccccccccc}
\toprule
& & \multicolumn{4}{c}{\textbf{PRISM}} & \multicolumn{4}{c}{\textbf{Community Alignment}} & \multicolumn{4}{c}{\textbf{Anthropic HH}} \\
& & \multicolumn{2}{c}{\textbf{Pythia-2.8B}} & \multicolumn{2}{c}{\textbf{Llama-3.2--1B}} &
      \multicolumn{2}{c}{\textbf{Pythia-2.8B}} & \multicolumn{2}{c}{\textbf{Llama-3.2--1B}} &
      \multicolumn{2}{c}{\textbf{Pythia-2.8B}} & \multicolumn{2}{c}{\textbf{Llama-3.2--1B}} \\
\textbf{Method} & \textbf{$\beta$ LR} &
\textbf{WR} & \textbf{LC-WR} &
\textbf{WR} & \textbf{LC-WR} &
\textbf{WR} & \textbf{LC-WR} &
\textbf{WR} & \textbf{LC-WR} &
\textbf{WR} & \textbf{LC-WR} &
\textbf{WR} & \textbf{LC-WR} \\
\midrule

DPO & -- &
51.09 & 50.42 &
64.96 & 64.33 &
35.85 & 36.87 &
70.69 & \textbf{66.53} &
69.29 & 69.34 &
71.58 & 71.08 \\

Lognormal (ours) & $10^{-4}$ &
59.61* & 58.28 &
67.56 & 66.73 &
41.18 & 41.83 &
\textbf{70.75} & 65.58 &
\textbf{71.65} & \textbf{72.01} &
\textbf{73.82} & \textbf{73.84} \\

Gamma (ours) & $10^{-4}$ &
\textbf{62.29}* & \textbf{61.39} &
\textbf{67.88} & \textbf{66.97} &
\textbf{47.64}* & \textbf{46.28} &
67.53 & 63.72 &
58.24* & 56.92 &
54.85* & 52.83 \\

\bottomrule
\end{tabular}}
\caption{
\textbf{Aggregate alignment performance across three datasets (PRISM, Community Alignment, Anthropic HH) and two base models (Pythia-2.8B, Llama-3.2-1B).}
We report win rate (WR) and length-controlled win rate (LC-WR). Statistical significance for WR is assessed using paired bootstrap confidence intervals and sign-flip permutation tests over prompt level preferences; LC-WR is reported without significance testing due to the absence of per prompt length-controlled annotations in AlpacaEval. MixDPO achieve aggregate performance that is in many cases higher than standard DPO, with the strongest gains appearing on datasets with greater preference heterogeneity (PRISM). These results illustrate that MixDPO can improve subgroup level alignment while maintaining competitive (and often improved) performance on aggregate metrics.
}
\label{tab:aggregate_three_datasets_heat}
\end{table*}

\section{Results}
\label{sec:results}

\subsection{Aggregate Metrics Mask Subgroup Level Alignment Differences}
\label{ssec:aggregated}

Figure~\ref{fig:education_micro_macro} compares subgroup level preference margins across annotator education levels for DPO trained Llama-1B model evaluated on the PRISM and Community Alignment datasets. Each point corresponds to the mean preference margin within an education subgroup. The blue horizontal line denotes the micro averaged margin, which weights each comparison equally, while the orange line denotes the macro averaged margin, which weights education subgroups equally.

Across both datasets, subgroup level alignment behavior varies substantially. Some education groups exhibit preference margins well above the aggregate micro average, while others fall far below it. Despite this dispersion, the averaged metric collapses these heterogeneous signals into a single value. Notably, underrepresented education groups (those appearing on the left of each panel) tend to exhibit lower preference margins than both the micro and macro averages. Because these groups contribute fewer comparisons, their degraded alignment is downweighted under micro averaging. Macro averaging partially corrects for this by assigning equal weight to each subgroup to better reflect subgroup level performance.

This pattern is consistent across PRISM and Community Alignment, indicating that the masking effect is not specific to a particular dataset or prompt distribution. These results demonstrate that aggregate metrics can mask meaningful subgroup level alignment differences.

\subsection{MixDPO Improves Subgroup Preference Preservation Without Aggregate Tradeoffs}
\label{ssec:h2}

We next evaluate whether modeling preference strength as a distribution improves the preservation of subgroup conditioned human preferences. We focus on the PRISM dataset, which is explicitly designed to capture heterogeneous and potentially conflicting value judgments across annotators, and on a $\beta$ learning rate of $10^{-4}$, which yields consistently strong aggregate performance (\S~\ref{app:winrates}). We focus on \emph{preference margins} since they are computed directly from the log likelihood assigned to observed human preference labels, and therefore provide a grounded signal of how well each method models annotated pairwise comparisons. 

Figure~\ref{fig:prism_subgroup_margins} reports preference margin gains over standard DPO for Pythia-2.8B. Each point shows the change in mean preference margin relative to DPO for a particular subgroup or aggregate statistic computed from held out human preference labels. Faded points denote individual demographic subgroups, while larger square markers denote macro and micro averaged margins. The figure is organized into three regions. From left to right, these correspond to \textbf{demographic subgroups} (annotator heterogeneity), \textbf{conversational framing} (contextual heterogeneity), and \textbf{aggregate averages}.

\paragraph{Demographic Heterogeneity.}
We first examine heterogeneity arising from differences across annotator demographics. Across all demographic dimensions, both our LogNormal and Gamma MixDPO achieve positive margin gains relative to DPO. These gains are reflected in the macro averaged margins, which indicates improved preservation of subgroup conditioned preference signals. Importantly, micro averaged margins also improve relative to DPO, indicating that subgroup gains do not come at the expense of average case performance.

\paragraph{Contextual Heterogeneity.}
Beyond demographic subgroups, we consider conversational framing as a source of subgroup structure. The middle panel of Figure~\ref{fig:prism_subgroup_margins} report margin gains disaggregated by conversational framing (Unguided, Values Guided, and Controversy Guided). Across prompt types, MixDPO yields positive margin gains over DPO.

\paragraph{Aggregate Performance.}
The right of Figure~\ref{fig:prism_subgroup_margins} reports macro and micro averaged preference margin gains across all subgroups. Both MixDPO variants improve macro averaged margins relative to DPO while preserving micro averaged margins.

While preference margins allow us to directly assess how well learned policies align with observed human preferences, they do not capture downstream generation quality. We therefore additionally report win rates to further assess aggregate alignment performance. Table~\ref{tab:aggregate_three_datasets_heat}, reports win rates across datasets and base models. MixDPO achieve aggregate win rates that are often higher than standard DPO across all three datasets and base models. On PRISM, both LogNormal and Gamma substantially improve win rate relative to DPO (for example, on Pythia-2.8B, LogNormal increases win rate from 51.1\% to 59.6\%, while Gamma reaches 62.3\%). On Anthropic HH, LogNormal consistently performs best.

These findings show that distributional modeling of preference strength improves subgroup preference without affecting aggregate performance. Full demographic level breakdowns are provided in Appendix~\ref{app:subgroup_results}.

\begin{figure}
    \centering
    \includegraphics[width=\linewidth]{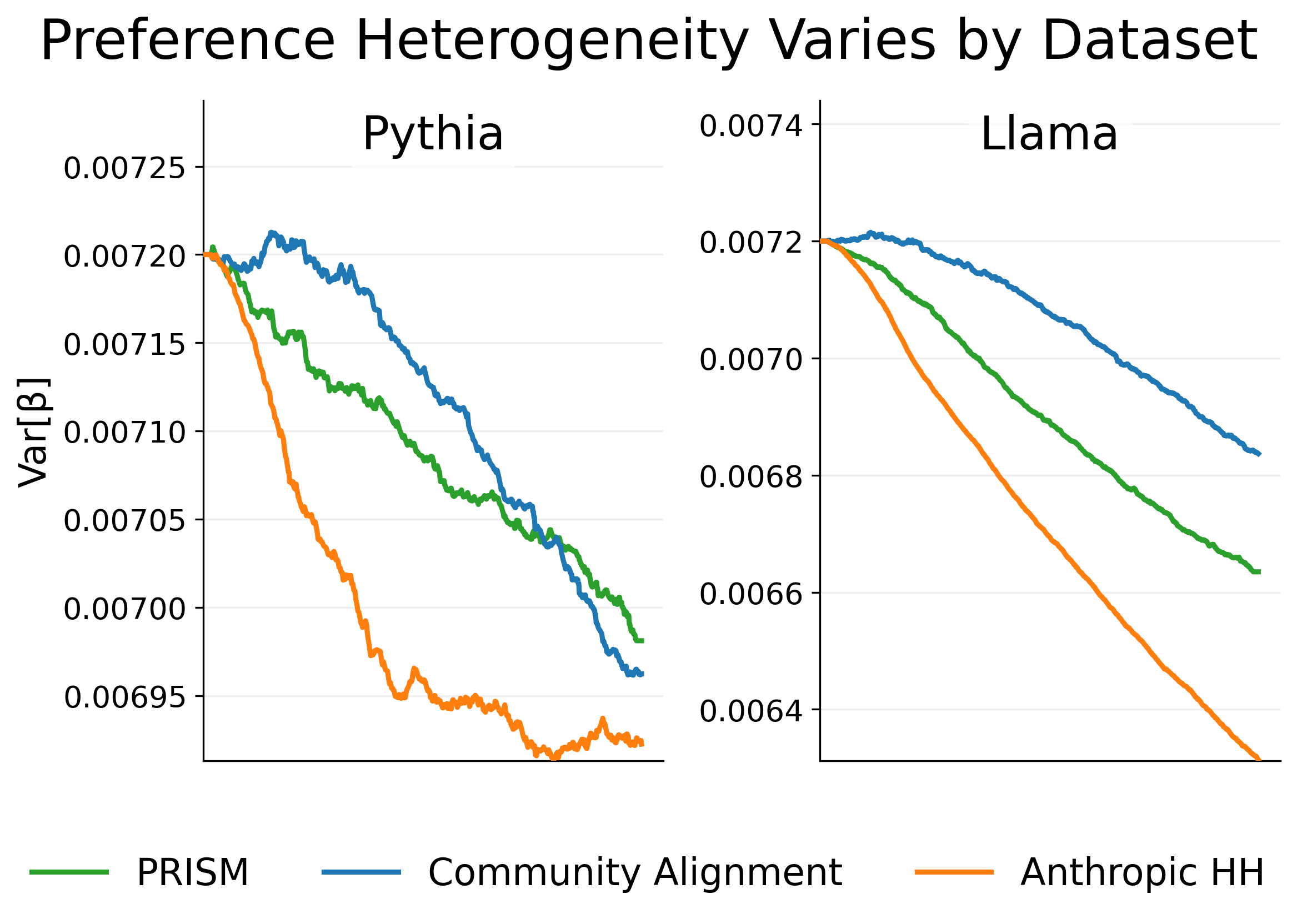}
    \caption{
          \textbf{Training trajectories of the variances of the learned preference strength distribution under Gamma MixDPO} (Pythia-2.8B, Llama-1B; $\beta$ learning rate $10^{-4}$). Anthropic HH converges to lower mean and variance, while PRISM and Community Alignment retain higher mean and variance throughout training. 
        }
    \label{fig:gamma_beta_trajectories}
\end{figure}

\subsection{MixDPO Makes Preference Heterogeneity in Datasets Measurable}
\label{ssec:distributional}

Beyond improving subgroup level alignment behavior, modeling preference strength as a distribution provides an insight into how preferences are expressed across different datasets. Unlike standard DPO, which fixes a single global preference strength, MixDPO enables analysis of the dispersion of preferences. 

We analyze the learned preference strength distributions across datasets by examining the variance of $\beta$. Differences in the learned variance across datasets reflect how much variation in effective preference strength is required to fit the observed preference judgments. This analysis highlights a benefit of distributional alignment where preference heterogeneity becomes an explicit and measurable quantity.

\begin{figure}
    \centering
    \includegraphics[width=\linewidth]{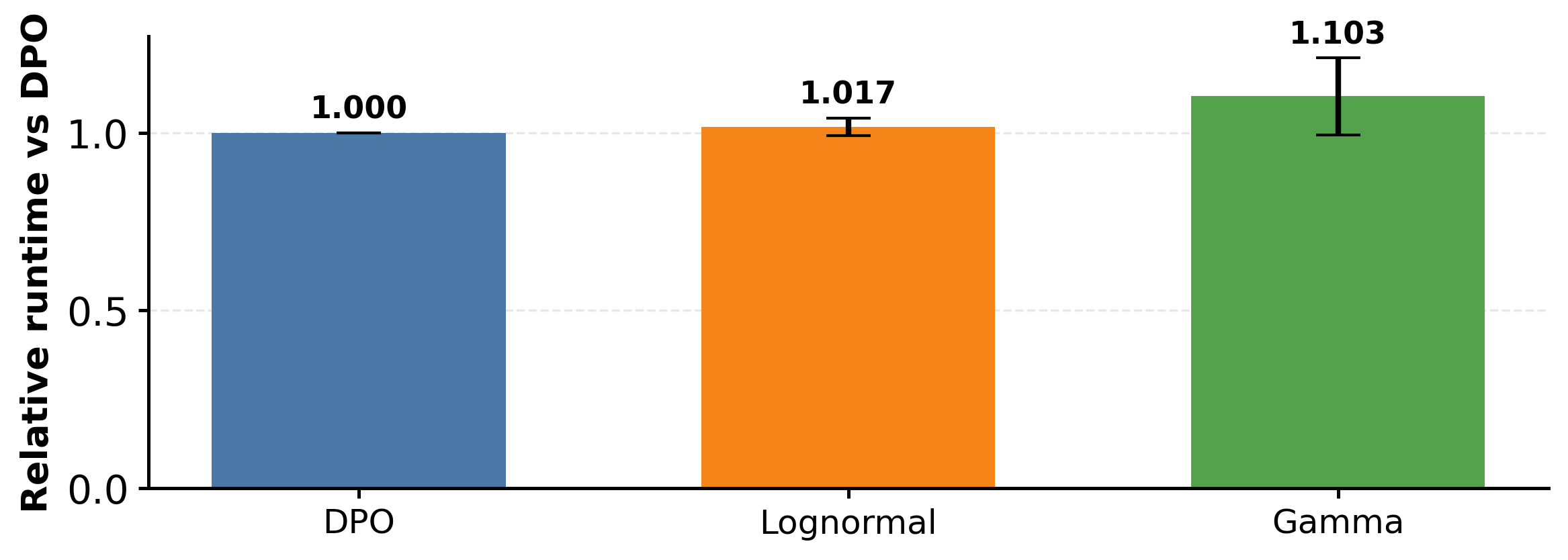}
    \caption{
        \textbf{Relative runtime comparison.} MixDPO introduces modest computational overhead: the LogNormal variant adds cost due to Monte Carlo sampling, while the Gamma variant incurs additional overhead from numerical evaluation of closed form expectation.
    }
    \label{fig:time}
\end{figure}

Figure~\ref{fig:gamma_beta_trajectories} shows the training trajectories of the variance of a Gamma distributed $\beta$ for Pythia-2.8B and Llama-1B. Despite identical training conditions, the learned distributions diverge across datasets. On Anthropic HH, the variance of $\beta$ decrease rapidly, which suggests that a limited range of effective preference strengths suffices to model the dataset. This is consistent with prior documentation that it is constructed from a smaller and more constrained annotator pool and focuses on a limited set of value dimensions (helpfulness and harmlessness; Appendix~\ref{app:dataset_heterogeneity}). In contrast, PRISM and Community Alignment exhibit slower decay in the variance of $\beta$. Because MixDPO does not condition on subgroup annotations, the observed differences in learned variance reflect dataset level structure encoded implicitly in the preference data. Across models, the learned variance exhibits a stable dataset level ordering, with Community Alignment generally highest and Anthropic HH tending toward lower variance, while the detailed dynamics depend on model and parameterization (Appendix~\ref{app:beta_trajectories}). These results support interpreting $\beta$ as a preference strength parameter that captures dataset level heterogeneity in how strongly preferences are expressed.

\subsection{Computational Overhead}
We evaluate the computational cost of MixDPO relative to standard DPO by measuring training time under identical hardware, batch size, sequence length, and optimization settings. To isolate algorithmic overhead from differences due to model size or dataset scale, we report relative runtime normalized to DPO and average across base models and datasets. Error bars denote $\pm 1$ standard deviation across settings.

Figure~\ref{fig:time} shows that MixDPO introduces modest overhead. The LogNormal variant incurs a small increase in runtime ($1.02 \times$), attributable to Monte Carlo sampling with $K=16$. The Gamma variant exhibits slightly higher overhead ($1.1 \times$), despite avoiding Monte Carlo estimation. This overhead arises from the numerical evaluation of special functions required by the closed form expectation.


\section{Discussion}
Our results first showed that micro averaged metrics downweight underrepresented subgroups, obscuring substantial variation in preference margins across demographics. Second, MixDPO preserves subgroup preferences without sacrificing aggregate performance where MixDPO improves macro averaged preference margins while maintaining or improving aggregate win rates.

\paragraph{When does distributional modeling help?}
The benefits of MixDPO are most prominent on datasets with preference heterogeneity. On PRISM, which captures value diversity across annotators, MixDPO improves both aggregate performance (+11.2 points win rate) and subgroup averaged preference margins. In contrast, on Anthropic HH, the learned $\beta$ distribution collapses to low variance, and MixDPO offers smaller gains. This suggests that distributional modeling learns broad distributions when heterogeneity is present and collapses to a narrower distribution when it is not.

\paragraph{Making preference heterogeneity measurable.}
Beyond performance improvements, MixDPO makes preference heterogeneity measurable. The learned variance of $\beta$ provide interpretable summaries of dataset level diversity. Datasets with consistent preferences induce low variance distributions, while datasets with ambiguous or conflicting judgments retain higher variance throughout training. This measurability provides insight into why alignment methods perform differently across datasets.

\paragraph{Implications for pluralistic alignment.}
Our findings suggest alignment under value pluralism that does not require explicit demographic conditioning. By modeling preference strength as a distribution, MixDPO enables the objective to model strong consensus preferences while respecting weaker or more conflicting judgments. This aligns with the goal of pluralistic alignment to not optimize for a fictitious average user but to preserve the distribution of values present in human feedback.
\section{Conclusion}
We introduce MixDPO, a generalization of DPO that models preference strength as a learned distribution rather than a fixed scalar. By capturing variation in how strongly preferences are expressed across training examples, MixDPO enables alignment objectives to preserve heterogeneous preference signals without sacrificing aggregate performance. This work establishes distributional preference modeling as a promising direction for pluralistic alignment, where the goal is not to optimize for a single average user, but to respect the diversity of values and perspectives present in human feedback data.

\section*{Limitations}

Unlike standard mixed logit models in economics, we did not condition utilities on observable attributes of the outputs or observable characteristics of raters. Instead, all output attributes are absorbed into a scalar reward model, and heterogeneity enters exclusively through a random coefficient capturing rater-specific sensitivity to this reward. This specification is appropriate for LLM alignment settings, where outputs are high-dimensional and rater attributes are often unavailable or only weakly informative. Our focus was therefore on unobserved heterogeneity in preference strength rather than systematic variation in preference direction. This approach can be interpreted as modeling disagreement in how strongly raters respond to a shared latent notion of quality, rather than disagreement over which dimensions constitute quality. An important limitation of this choice is that it abstracts away from potentially meaningful structure in observable output features or rater characteristics; incorporating such conditioning into the reward or preference model is a natural direction for future work.

We evaluate generation quality using AlpacaEval with GPT-4-Turbo as a judge rather than human evaluations. While prior work shows strong correlation with human judgments, LLM-based evaluation may introduce biases that we cannot fully control. To partially mitigate this concern, we additionally evaluate learned policies using the log likelihood of observed preference data, which is directly derived from human annotated pairwise comparisons in the alignment datasets. This provides a grounded signal of how well each method assigns probability to observed human preference labels.
Nevertheless, such likelihood based analyses cannot fully substitute for human judgments of generation quality. Conducting dedicated human evaluations to further validate these findings remains an important direction for future work.





\bibliography{custom}

\appendix
\onecolumn
\section{Background on Mixed Logit Models}
\label{app:mixed_logit}
\paragraph{Historical motivation and IIA}
The mixed logit (also known as the random-coefficients logit) model was introduced to overcome fundamental limitations of the standard multinomial logit, most notably the independence of irrelevant alternatives (IIA) property and the assumption of homogeneous preferences across decision makers \cite{McFadden1974}. In many applications—transportation choice, marketing, health, and policy evaluation—agents differ systematically in their sensitivities to attributes such as price, time, risk, or quality, and these differences cannot be adequately captured by fixed coefficients. Mixed logit addresses this limitation by allowing coefficients to vary randomly across individuals according to a specified distribution, thereby generating flexible substitution patterns and accommodating rich forms of unobserved heterogeneity. The model’s modern formulation and its approximation properties were formalized by \cite{McFaddenTrain2000}, who showed that mixed logit can approximate any random utility model arbitrarily well under mild regularity conditions, and were subsequently popularized in applied work by \cite{train2009}.

\paragraph{Distributional choices and estimation}
A central modeling choice in mixed logit concerns the specification of the distribution of random coefficients. Common parametric choices include normal distributions, which allow coefficients to take both positive and negative values; log-normal distributions, often used to impose sign restrictions such as strictly negative price sensitivity \cite{train2009}; and truncated or bounded distributions when economic theory dictates strict constraints \cite{Hensher_Rose_Greene_2005}. More flexible specifications—such as mixtures of normals or latent-class (finite-mixture) formulations—have been employed to capture multimodality and discrete segments in preferences \cite{McLachlanPeel2000, train2008algorithms}. While richer distributions improve behavioral realism, they introduce tradeoffs between interpretability, computational burden, and identification, particularly in high-dimensional settings. Estimation is typically conducted via simulated maximum likelihood or Bayesian methods, relying on Monte Carlo or quasi-Monte Carlo integration to approximate choice probabilities \cite{train2009}.

\paragraph{Prior Applications}
Mixed logit models have been used successfully across a wide range of domains, including transportation mode choice \cite{Bhat2001}, product demand estimation \cite{BerryLevinsohnPakes1995}, environmental valuation \cite{TrainWeeks2005}, health-care decision making \cite{Hensher_Rose_Greene_2005}, and policy analysis, where accounting for heterogeneous responses is essential for credible counterfactual prediction and welfare analysis. 

\section{Monte Carlo Approximation and Reparameterization with Lognormal Preference Strength}
\label{app:mc_reparam}
This appendix provides implementation and optimization details for the LogNormal preference strength distribution introduced in Section~\ref{sec:mldpo}. We focus on the Monte Carlo approximation, parameterization choices, and the resulting training objective.

\paragraph{Monte Carlo Approximation}
For distributions that do not admit a closed-form marginalization of
$\mathbb{E}_{\beta}[\sigma(\beta \Delta r_\theta)]$, we approximate the expectation using Monte Carlo sampling:
\begin{equation}\small
\mathbb{E}_{\beta_0 \sim p(\beta_0)} \sigma\!(\beta ( \Delta r_\theta )) \approx \frac{1}{K} \sum_{k=1}^{K} \sigma (\beta_0^{(k)} \Delta r_\theta)
\end{equation} where $K=16$ samples are used per training update.

\paragraph{Parameterization of the LogNormal Distribution.}
We parameterize the beta distribution $p(\beta_0)$ using location and scale parameters $\theta_{\beta} = (\mu, \sigma)$, where $\mu \in \mathbb{R}$ and $\sigma > 0$. To guarantee $\sigma > 0$, we use the softplus transformation $\sigma = softplus(\tilde{\sigma}) = \log (1 + \exp(\tilde{\sigma}))$ \cite{dugas2000incorporating}.

\paragraph{Reparameterized Sampling.}
We use the reparameterization trick \cite{Kingma14, pmlr-v32-rezende14} to enable end-to-end optimization of both the policy parameters $\theta$ and the heterogeneity parameters $\theta_\beta$. Instead of sampling $\beta_0$ directly from $p(\beta_0)$, we express it as a deterministic and differentiable transformation of a noise variable $\varepsilon \sim \mathcal{N}(0,1)$, which is $\beta_0 = \exp(\mu + \sigma \varepsilon)$. This reparameterization makes the sampling process differentiable with respect to $(\mu, \sigma)$, and allows gradients to propagate through $\beta_0$ to optimize $\theta_\beta$ using backpropagation. This reparameterized gradient has lower variance than score function estimators such as the log derivative trick \cite{xu2019variance}.

\paragraph{Resulting Training Objective.}
Under this parameterization, the MixDPO loss optimized becomes
\begin{equation}
\mathcal{L}_{\text{ML-DPO}}(\theta, \theta_\beta) = 
 -\mathbb{E}_{(x,y_w,y_l), \epsilon \sim \mathcal{N}(0,1)} [
\log \frac{1}{K} \sum_{k=1}^{K} \sigma(\exp(\mu + \sigma \epsilon_k) \Delta r_\theta)]
\end{equation}
This reparameterized Monte Carlo estimator yields unbiased gradients but introduces variance through sampling, in contrast to the analytic Gamma formulation described in Appendix~\ref{app:gamma_closedform}.

\section{Closed Form Inner Expectation with Gamma Prior}
\label{app:gamma_closedform}
This appendix provides the derivation and implementation details for the closed form Gamma preference strength distribution presented in Section~\ref{sec:mldpo}. We derive the analytic expressions for
$\mathbb{E}_{\beta}[\sigma(\beta \Delta r_\theta)]$
and discuss how the resulting form enables end-to-end optimization of the preference strength distribution.

\paragraph{Derivation of the closed form with Gamma prior}
\label{app:closedform_derivation}
Let $\beta_0 \sim \Gamma(k,\lambda)$ (shape $k>0$, rate $\lambda>0$) with pdf
$f(\beta)=\frac{\lambda^k}{\Gamma(k)}\beta^{k-1}e^{-\lambda\beta}$, and let
$\Delta r_\theta = r_\theta(x, y_w) - r_\theta(x, y_l)$. We want to compute: 
\begin{equation}
    \mathbb{E}_{\beta_0 \sim \Gamma(\beta_0)}
    \sigma(\beta ( \Delta r_\theta )) = \int_0^\infty \frac{1}{1+e^{-\Delta r_\theta \beta}} \frac{\lambda^k}{\Gamma(k)} \beta^{k-1} e^{-\lambda \beta} d \beta
\end{equation}
using the fact that (i) For $z>0$,
$\frac{1}{1+e^{-z}}=1+\sum_{n=1}^{\infty}(-1)^n e^{-n z}$
(ii) Gamma integral: $\int_0^\infty \beta^{k-1}e^{-a\beta}d\beta=\Gamma(k)a^{-k}$ $(a>0)$
(iii) Lerch transcendent function: $\Phi(z,s,a)=\sum_{n=1}^\infty \frac{z^n}{(a + n)^s}$

\noindent\paragraph{Case 1:} $\Delta r_\theta > 0$.
Use (i) with $z=\Delta r_\theta \beta$:
\begin{equation}
    \frac{1}{1+e^{-\Delta r_\theta \beta}} = 1+\sum_{n=1}^\infty (-1)^n e^{-n\Delta r_\theta \beta}
\end{equation}
\begin{align}
\mathbb{E}_{\beta_0 \sim \Gamma(\beta_0)}
\sigma(\beta \Delta r_\theta)
&=
\int_0^\infty \Big(1 + \sum_{n=1}^\infty (-1)^n e^{-n\Delta r_\theta \beta}\Big)
\frac{\lambda^k}{\Gamma(k)} \beta^{k-1} e^{-\lambda \beta} d\beta
\\
&= 
1 + \sum_{n=1}^\infty (-1)^n \frac{\lambda^k}{\Gamma(k)}
\int_0^\infty \beta^{k-1} e^{-(\lambda+n\Delta r_\theta )\beta} d\beta
\end{align}
Apply (ii) where $\alpha = \lambda + n\Delta r_\theta$:
\begin{equation}
    1 + \sum_{n=1}^\infty (-1)^n \frac{\lambda^k}{(\lambda+n\Delta  r_\theta)^k}
    = 1 + \Big(\frac{\lambda}{\Delta  r_\theta}\Big)^{\!k} \sum_{n=1}^\infty
    \frac{(-1)^n}{(\lambda/\Delta  r_\theta + n)^k}
\end{equation}
Applying (iii) where $z = -1, s = n, a = \frac{\lambda}{\Delta r_\theta}$:
\begin{equation}
    \mathbb{E}_{\beta_0 \sim \Gamma(\beta_0)} \sigma(\beta \Delta r_\theta) = 1 + \Big(\frac{\lambda}{\Delta  r_\theta}\Big)^{\!k} \Phi(-1, n, \frac{\lambda}{\Delta r_\theta})
\end{equation}

\noindent\paragraph{Case 2:} $\Delta r_\theta < 0$. Let $t=-\Delta r_\theta>0$ and use $\sigma(-z)=1-\sigma(z)$:
\begin{equation}
    \mathbb{E}_{\beta_0 \sim \Gamma(\beta_0)}[\sigma(-t\beta_0)] = 1 - \mathbb{E}_{\beta_0 \sim \Gamma(\beta_0)}[\sigma(t\beta_0)]
\end{equation}
Applying the $\Delta r_\theta > 0$ result:
\begin{equation}
    \mathbb{E}_{\beta_0 \sim \Gamma(\beta_0)} \sigma(\beta \Delta r_\theta) = 1 - \Big[1 - \Big(\frac{\lambda}{t}\Big)^{\!k} \Phi\!\Big(-1,k,1+\frac{\lambda}{t}\Big)\Big] = \Big(\frac{\lambda}{-\Delta r_\theta}\Big)^{\!k} \Phi\!\Big(-1,k,1+\frac{\lambda}{-\Delta r_\theta}\Big)
\end{equation}

\noindent\paragraph{Case 3:} $\Delta r_\theta = 0$. 
\begin{equation}
    \mathbb{E}_{\beta_0 \sim \Gamma(\beta_0)} \sigma(\beta \Delta r_\theta) = \mathbb{E}_{\beta_0 \sim \Gamma(\beta_0)} \sigma(\beta \cdot 0) = \frac{1}{2}
\end{equation}

\subsection{Numerical Evaluation of Special Functions}
\label{app:gamma_numerics}
We approximate the Hurwitz zeta function using a truncated series expansion with $n=1000$ terms. Empirically, this truncation yields an absolute error of approximately $2.1 \times 10^{-5}$ while incurring negligible computational overhead (on the order of microseconds per evaluation). Since the approximation is expressed as a finite sum of differentiable operations, gradients propagate through the resulting objective via automatic differentiation.

\section{Dataset Heterogeneity and Annotator Diversity}
\label{app:dataset_heterogeneity}
In this appendix, we provide additional context for our characterization of
datasets in terms of \emph{observable preference heterogeneity}, as referenced
in \S\ref{ssec:datasets}. 

\paragraph{Annotator Diversity.}
One source of observable heterogeneity arises from the size and composition of the annotator pool. PRISM and Community Alignment were explicitly designed to increase annotator diversity through broad recruitment and the collection of auxiliary metadata. PRISM includes over 1{,}500 annotators with associated demographic attributes and value profiles \cite{kirk2024prism}. Community Alignment contains preference judgments from 3{,}196 unique annotators prior to English filtering, spanning multiple cultural and linguistic contexts \cite{zhang2025cultivating}.

In contrast, Anthropic HH was constructed using crowd workers recruited through Amazon Mechanical Turk. According to \cite{bai2022training} (p.~68), the dataset relies on a pool of approximately 115 unique annotators, and does not release per sample annotator metadata. The absence of disaggregated metadata prevents direct analysis of preference variation across demographic groups. As a result, annotator level heterogeneity is less observable in Anthropic HH than in PRISM and Community Alignment.

\paragraph{Contextual and Value Diversity.}
A second source of observable heterogeneity arises from the diversity of prompts and the range of values they are designed to elicit. PRISM and Community Alignment include prompts that explicitly encourage value conflict, disagreement, or context dependent judgment. In PRISM, conversational framing (\textit{Unguided}, \textit{Values Guided}, \textit{Controversy}) induces variation in how preferences are expressed, which we treat as a form of contextual subgroup in our analysis. Community Alignment similarly targets value laden judgments grounded in established cultural value dimensions.

By comparison, Anthropic HH focuses on helpfulness and harmlessness, which is a more constrained set of values. While this design is well suited for evaluating safety and helpfulness oriented alignment, it induces less variation in the types of values.

\paragraph{Interpretation.}
These differences motivate our treatment of Anthropic HH as a
\emph{relatively preference homogeneous control case} with respect to observable
annotator and context level variation.

\section{Training Details}
\label{app:training_details}

\paragraph{Models and Initialization.}
All experiments are conducted using open-weight language models on HuggingFace. We report results for Pythia-2.8B and Llama-1B. For each dataset, models are initialized from a supervised fine-tuned (SFT) checkpoint trained on the same preference dataset, and alignment training proceeds by further optimizing the policy using DPO or MixDPO objectives. The reference model is fixed throughout training and uses \texttt{float32} precision.

\paragraph{Optimization and Precision.}
We train using RMSProp with learning rate $5\times10^{-7}$ for policy parameters. All models are trained with Fully Sharded Data Parallel (FSDP) using mixed precision: policy parameters are stored in \texttt{bfloat16}, while reference model parameters remain in \texttt{float32}. Gradient clipping is applied with a maximum norm of 1.0. We do not use gradient accumulation.

\paragraph{Batching and Sequence Lengths.}
Unless otherwise specified, we use a batch size of 64 for training and 32 for evaluation. The maximum sequence length is 128 tokens, with a maximum prompt length of 64 tokens. Evaluation is performed every 1{,}984 updates on a held out validation split.

\paragraph{Training Duration.}
All runs are trained for a single epoch over the preference dataset. Each run processes approximately 55k preference comparisons for Anthropic HH, Community Alignment, and PRISM.

\section{Complete Win Rate Results}
\label{app:winrates}

This appendix reports complete win rate (WR) and length-controlled win rate (LC-WR) results across all datasets, base models, and $\beta$ learning rates considered in our study. These results provide context for the experimental choices made in \S~\ref{ssec:h2}, in particular our focus on a $\beta$ learning rate of $10^{-4}$.

Across both base models and all three datasets, a $\beta$ learning rate of $10^{-4}$ consistently yields strong and stable aggregate performance, avoiding the performance degradation observed at higher learning rates and the underfitting seen at lower learning rates. This trend holds for both lognormal and gamma parameterizations, which makes $\beta$ LR = $10^{-4}$ a robust learning rate across model families (Table~\ref{tab:aggregate_three_datasets_heat}).

\begin{table*}[t]
\centering
\setlength{\tabcolsep}{3.5pt}
\small
\resizebox{\textwidth}{!}{
\begin{tabular}{llcccccccccccc}
\toprule
& & \multicolumn{4}{c}{\textbf{PRISM}} & \multicolumn{4}{c}{\textbf{Community Alignment}} & \multicolumn{4}{c}{\textbf{Anthropic HH}} \\
& & \multicolumn{2}{c}{\textbf{Pythia-2.8B}} & \multicolumn{2}{c}{\textbf{Llama-3.2--1B}} &
      \multicolumn{2}{c}{\textbf{Pythia-2.8B}} & \multicolumn{2}{c}{\textbf{Llama-3.2--1B}} &
      \multicolumn{2}{c}{\textbf{Pythia-2.8B}} & \multicolumn{2}{c}{\textbf{Llama-3.2--1B}} \\
\textbf{Method} & \textbf{$\beta$ LR} &
\textbf{WR} & \textbf{LC-WR} &
\textbf{WR} & \textbf{LC-WR} &
\textbf{WR} & \textbf{LC-WR} &
\textbf{WR} & \textbf{LC-WR} &
\textbf{WR} & \textbf{LC-WR} &
\textbf{WR} & \textbf{LC-WR} \\
\midrule
DPO & -- &
\cellcolor[HTML]{4D99CA}58.67 & \cellcolor[HTML]{66ABD4}55.57 &
\cellcolor[HTML]{2C7CBA}65.89 & \cellcolor[HTML]{2B7BBA}64.20 &
\cellcolor[HTML]{F7FBFF}38.00 & \cellcolor[HTML]{F7FBFF}28.76 &
\cellcolor[HTML]{A6CBE3}60.61 & \cellcolor[HTML]{BFD7EA}54.39 &
\cellcolor[HTML]{08519C}69.63 & \cellcolor[HTML]{08519C}68.44 &
\cellcolor[HTML]{2C7CBA}75.33 & \cellcolor[HTML]{2C7CBA}75.18 \\
Lognormal & -- &
\cellcolor[HTML]{1C6BB0}63.50 & \cellcolor[HTML]{1A68AE}62.10 &
\cellcolor[HTML]{B7D3E8}61.74 & \cellcolor[HTML]{BDD7EC}59.41 &
\cellcolor[HTML]{DBEAF6}40.00 & \cellcolor[HTML]{D2E4F3}32.02 &
\cellcolor[HTML]{2B7BBA}67.68 & \cellcolor[HTML]{2B7BBA}62.02 &
\cellcolor[HTML]{084E98}\textbf{\cellcolor[HTML]{084E98}70.48} & \cellcolor[HTML]{084E98}\textbf{\cellcolor[HTML]{084E98}69.54} &
\cellcolor[HTML]{A6CBE3}71.00 & \cellcolor[HTML]{A6CBE3}70.86 \\
Lognormal & $10^{-5}$ &
\cellcolor[HTML]{3D86C0}61.00 & \cellcolor[HTML]{4B94C8}57.99 &
\cellcolor[HTML]{4D99CA}\cellcolor[HTML]{4D99CA}66.89 & \cellcolor[HTML]{4D99CA}\cellcolor[HTML]{4D99CA}65.37 &
\cellcolor[HTML]{EAF3FB}39.17 & \cellcolor[HTML]{EEF6FC}29.34 &
\cellcolor[HTML]{4D99CA}65.32 & \cellcolor[HTML]{66ABD4}60.06 &
\cellcolor[HTML]{2C7CBA}67.51 & \cellcolor[HTML]{2C7CBA}66.24 &
\cellcolor[HTML]{66ABD4}73.17 & \cellcolor[HTML]{66ABD4}72.98 \\
Lognormal & $10^{-4}$ &
\cellcolor[HTML]{3D86C0}61.00 & \cellcolor[HTML]{3A83BE}59.59 &
\cellcolor[HTML]{2B7BBA}65.67 & \cellcolor[HTML]{2C7CBA}64.04 &
\cellcolor[HTML]{A6CBE3}43.33 & \cellcolor[HTML]{A6CBE3}35.90 &
\textbf{\cellcolor[HTML]{084E98}69.02} & \cellcolor[HTML]{2B7BBA}\textbf{\cellcolor[HTML]{2B7BBA}62.99} &
\cellcolor[HTML]{66ABD4}65.32 & \cellcolor[HTML]{66ABD4}64.15 &
\textbf{\cellcolor[HTML]{084E98}76.83} & \textbf{\cellcolor[HTML]{084E98}76.78} \\
Lognormal & $10^{-3}$ &
\textbf{\cellcolor[HTML]{084E98}65.00} & \textbf{\cellcolor[HTML]{084E98}64.00} &
\cellcolor[HTML]{8DBBE0}64.00 & \cellcolor[HTML]{8DBBE0}61.84 &
\cellcolor[HTML]{2C7CBA}45.33 & \cellcolor[HTML]{2C7CBA}38.11 &
\cellcolor[HTML]{1A68AE}68.69 & \cellcolor[HTML]{8DBBE0}61.30 &
\cellcolor[HTML]{A6CBE3}63.80 & \cellcolor[HTML]{A6CBE3}63.66 &
\cellcolor[HTML]{2C7CBA}75.83 & \cellcolor[HTML]{2C7CBA}75.78 \\
Gamma & -- &
\cellcolor[HTML]{F7FBFF}49.67 & \cellcolor[HTML]{F7FBFF}45.93 &
\textbf{\cellcolor[HTML]{084E98}69.33} & \textbf{\cellcolor[HTML]{084E98}68.60} &
\cellcolor[HTML]{D2E4F3}40.13 & \cellcolor[HTML]{C7DBEF}33.00 &
\cellcolor[HTML]{66ABD4}65.66 & \cellcolor[HTML]{A6CBE3}58.11 &
\cellcolor[HTML]{F7FBFF}47.29 & \cellcolor[HTML]{F7FBFF}42.59 &
\cellcolor[HTML]{DBEAF6}61.70 & \cellcolor[HTML]{DBEAF6}59.89 \\
Gamma & $10^{-5}$ &
\cellcolor[HTML]{66ABD4}58.67 & \cellcolor[HTML]{66ABD4}55.44 &
\cellcolor[HTML]{C7DBEF}63.67 & \cellcolor[HTML]{D2E4F3}61.12 &
\cellcolor[HTML]{F7FBFF}37.50 & \cellcolor[HTML]{F7FBFF}28.41 &
\cellcolor[HTML]{BFD7EA}61.62 & \cellcolor[HTML]{D2E4F3}56.28 &
\cellcolor[HTML]{C7DBEF}57.38 & \cellcolor[HTML]{C7DBEF}54.68 &
\cellcolor[HTML]{F7FBFF}55.44 & \cellcolor[HTML]{F7FBFF}52.25 \\
Gamma & $10^{-4}$ &
\cellcolor[HTML]{145DA0}64.00 & \cellcolor[HTML]{145DA0}62.40 &
\cellcolor[HTML]{1C6BB0}68.33 & \cellcolor[HTML]{1A68AE}66.67 &
\textbf{\cellcolor[HTML]{084E98}47.00} & \textbf{\cellcolor[HTML]{084E98}41.53} &
\cellcolor[HTML]{A6CBE3}60.61 & \cellcolor[HTML]{BFD7EA}55.16 &
\cellcolor[HTML]{BFD7EA}58.08 & \cellcolor[HTML]{BFD7EA}53.50 &
\cellcolor[HTML]{DBEAF6}61.42 & \cellcolor[HTML]{DBEAF6}59.12 \\
Gamma & $10^{-3}$ &
\cellcolor[HTML]{2B7BBA}61.95 & \cellcolor[HTML]{2C7CBA}59.32 &
\cellcolor[HTML]{145DA0}69.00 & \cellcolor[HTML]{145DA0}67.68 &
\cellcolor[HTML]{A6CBE3}42.81 & \cellcolor[HTML]{A6CBE3}34.70 &
\cellcolor[HTML]{2C7CBA}67.00 & \cellcolor[HTML]{2C7CBA}62.63 &
\cellcolor[HTML]{C7DBEF}57.74 & \cellcolor[HTML]{C7DBEF}54.10 &
\cellcolor[HTML]{F7FBFF}45.81 & \cellcolor[HTML]{F7FBFF}45.95 \\
\bottomrule
\end{tabular}}
\caption{
Aggregate alignment performance across three datasets (PRISM, Community Alignment, Anthropic HH) and two base models (Pythia-2.8B, Llama3.2-1B). Standard DPO and lognormal with fixed beta parameters performs strongly on Anthropic HH, whereas gamma based MLDPO variants are competitive on PRISM and Community Alignment.}
\label{tab:aggregate_three_datasets_heat}
\end{table*}




\section{Additional Subgroup Results}
\label{app:subgroup_results}

\begin{figure*}
    \centering
    \includegraphics[width=\linewidth]{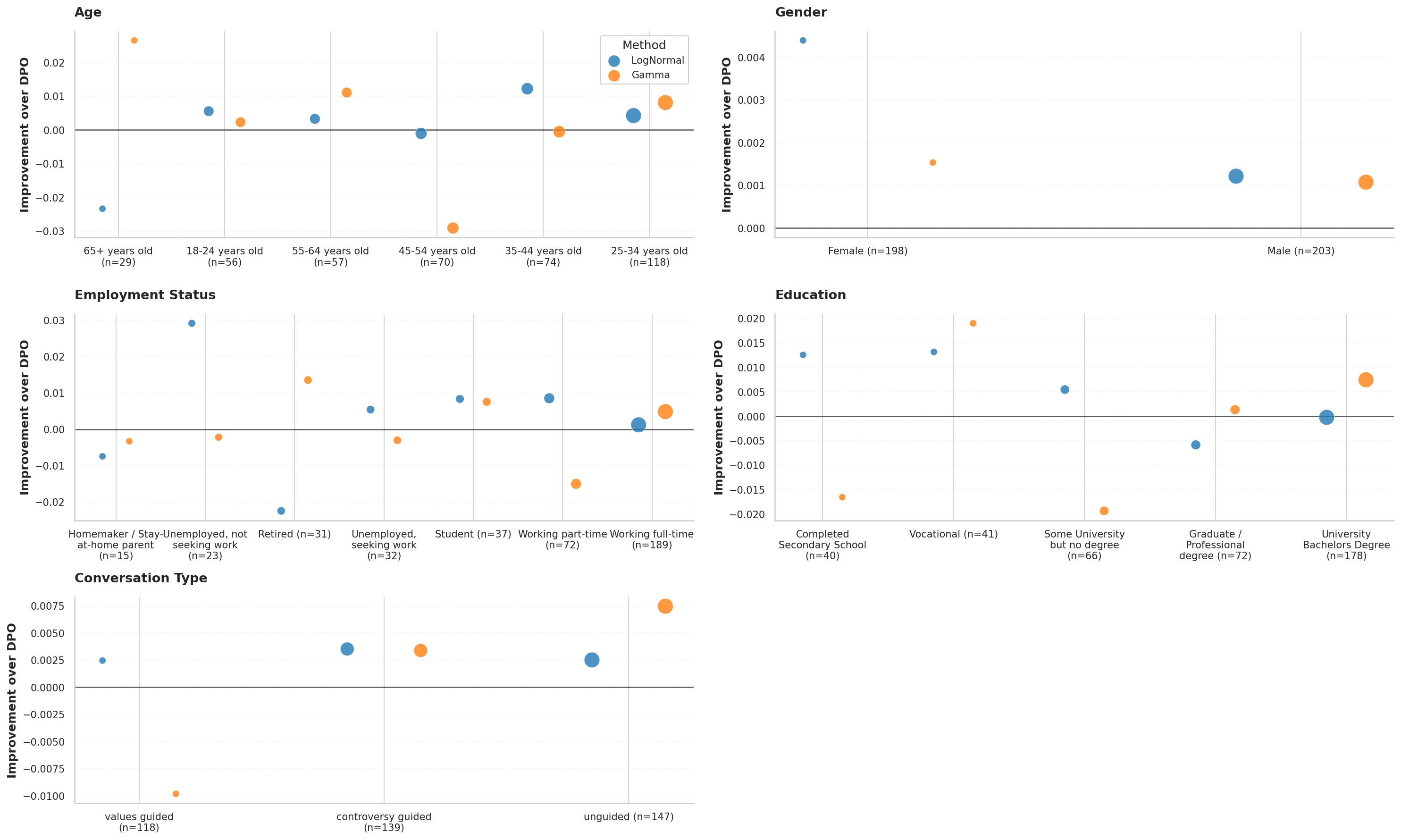}
    \caption{Preference margin gains over standard DPO on PRISM, disaggregated by annotator demographics and conversational framing. We report changes in mean preference margin relative to DPO for Pythia-2.8B. Points correspond to individual subgroups, with marker size proportional to the number of annotated comparisons. Colors indicate MixDPO variants (LogNormal, Gamma). Preference margins are shown for demographic attributes (Age, Gender, Employment Status, Education) and for conversational framing (Values Guided, Controversy Guided, Unguided). Positive values indicate improved alignment with observed human preferences relative to DPO.
}
    \label{fig:prism_subgroup_margins}
\end{figure*}

Figure~\ref{fig:prism_subgroup_margins} provides a complete breakdown of preference margin gains relative to DPO across annotator demographics and conversational framing on the PRISM dataset. Each panel reports the change in mean preference margin for a specific subgroup.

Across demographic dimensions and conversational framing, both LogNormal and Gamma MixDPO exhibit nonnegative margin gains relative to DPO. Marker size reflects subgroup sample size, illustrating that gains are observed across both large and small subgroups.

This figure complements the results reported in the main text by showing that improvements from distributional modeling of preference strength are not driven by a narrow subset of annotators or prompt types, but instead appear broadly across demographic and contextual subgroups.

\section{Additional Preference Strength Trajectories Across Base Models}
\label{app:beta_trajectories}
\begin{figure*}[t]
\centering
\begin{subfigure}[t]{0.31\textwidth}
  \vspace{0pt} 
  \centering
  \includegraphics[width=\textwidth]{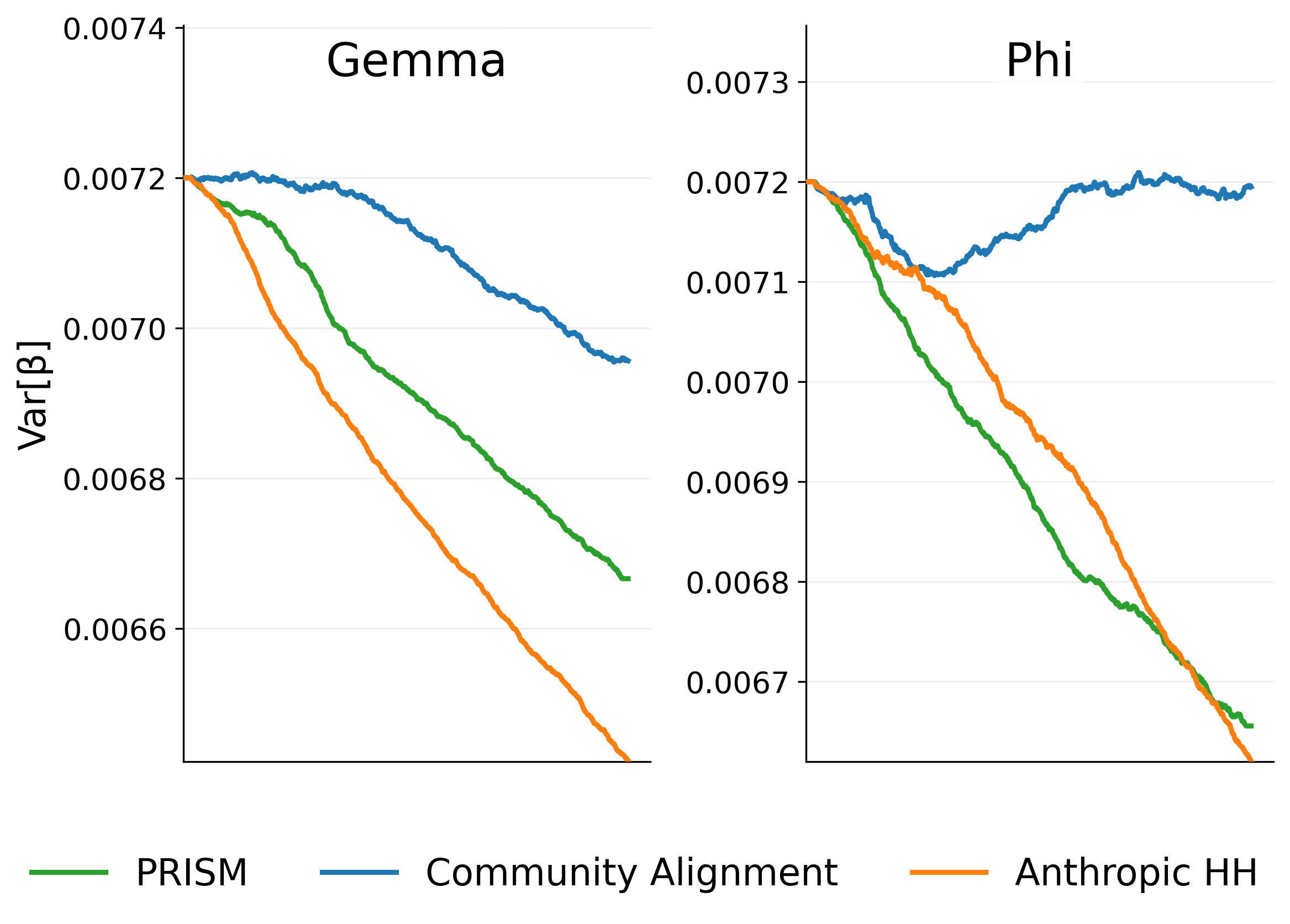}
  \caption{Gamma parameterization}
  \label{fig:beta_gamma}
\end{subfigure}
\hfill
\begin{subfigure}[t]{0.64\textwidth}
  \vspace{0pt} 
  \centering
  \begin{minipage}{0.49\textwidth}
    \centering
    \includegraphics[width=\textwidth]{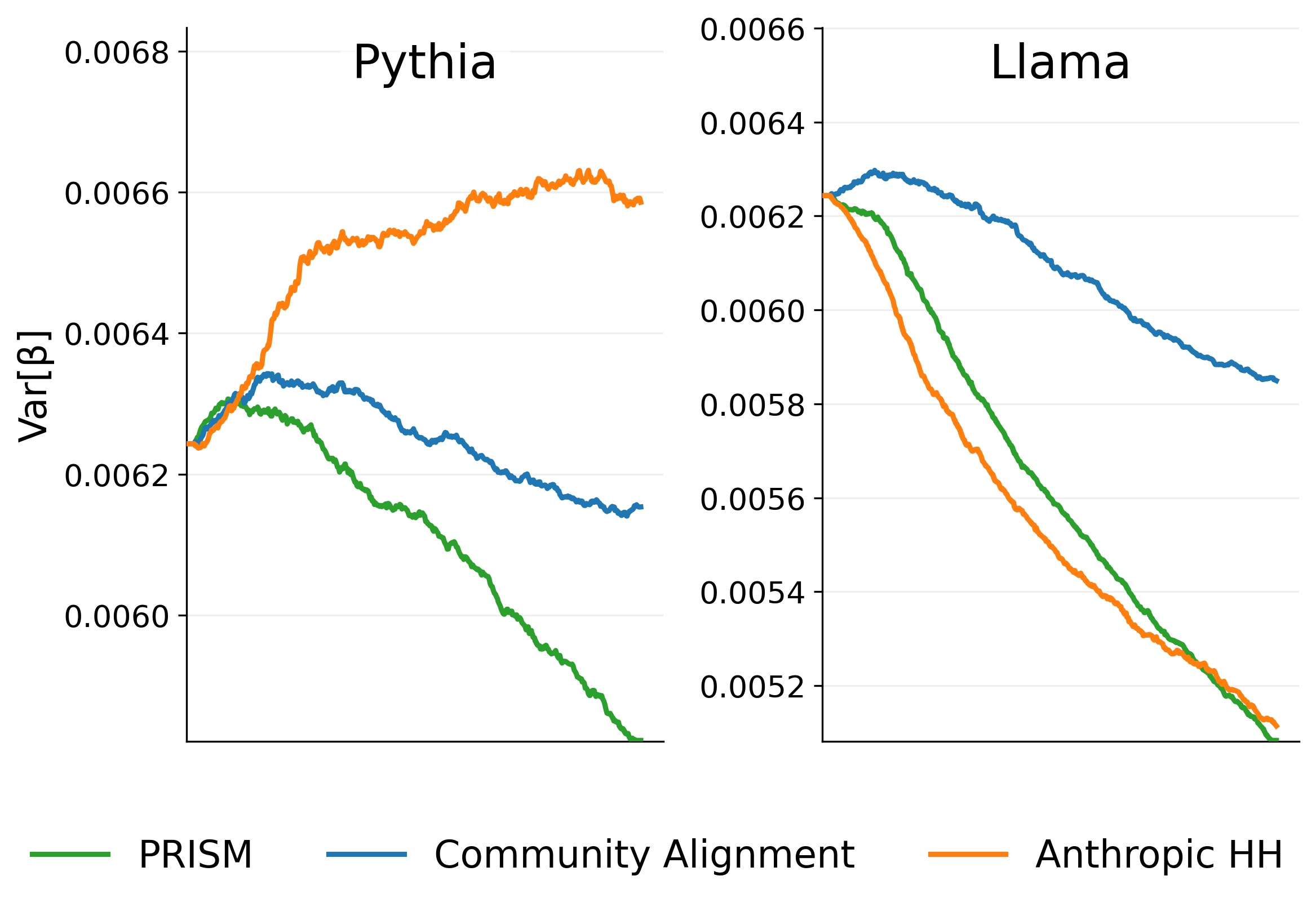}
  \end{minipage}
  \hfill
  \begin{minipage}{0.49\textwidth}
    \centering
    \includegraphics[width=\textwidth]{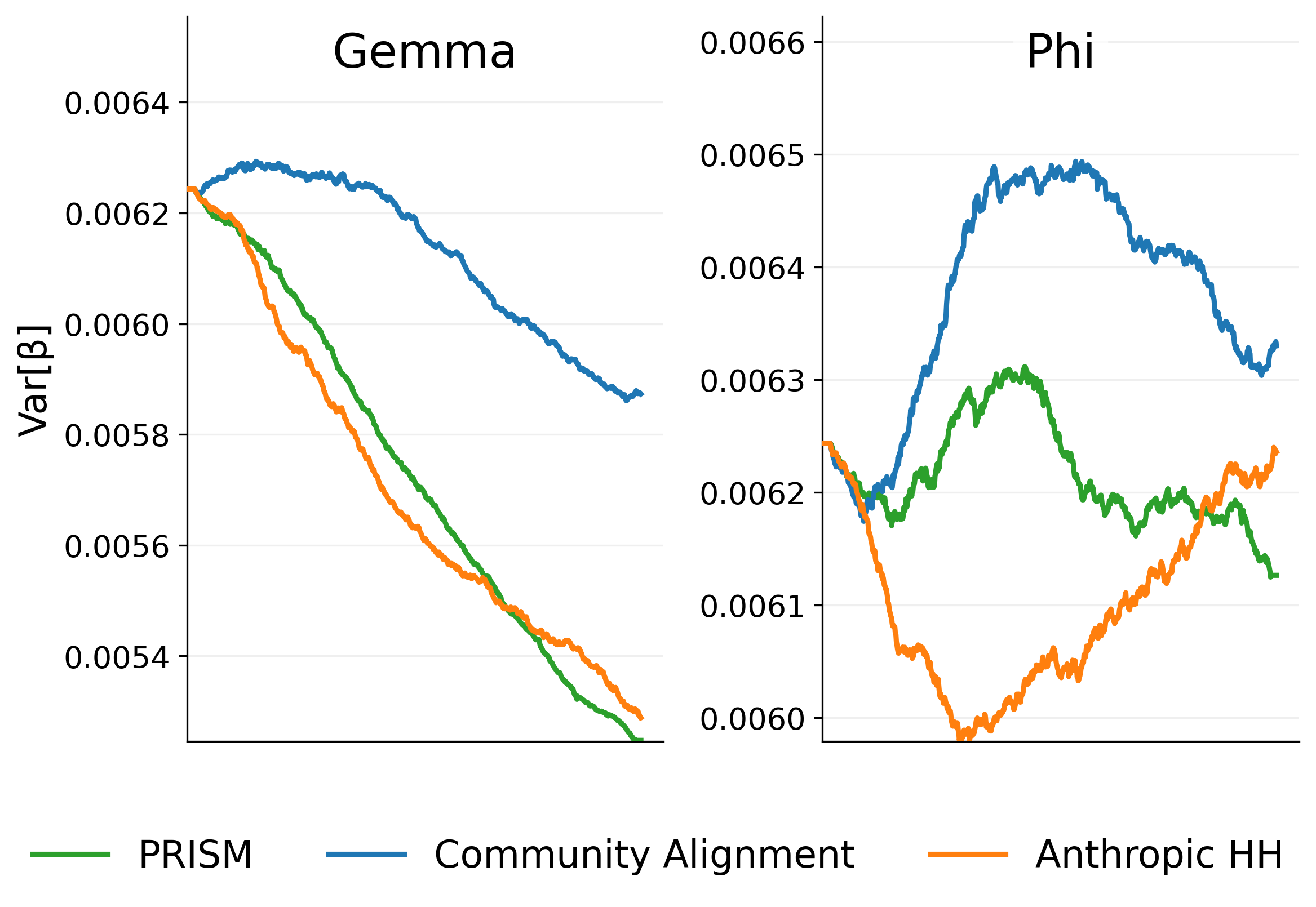}
  \end{minipage}
  \caption{LogNormal parameterization}
  \label{fig:beta_lognormal}
\end{subfigure}

\caption{
Training trajectories of the variance of the learned preference strength parameter $\beta$ under MixDPO across datasets and base models. Panel (a) shows results under two Gamma parameterization cases. Panel (b) shows four LogNormal cases. Across both parameterizations, trajectories exhibit a stable dataset level ordering: Community Alignment generally exhibits higher variance, PRISM occupies an intermediate regime, and Anthropic HH tends toward lower variance, although the precise dynamics depend on the model and distributional choice.}
\label{fig:beta_trajectories}
\end{figure*}

This appendix provides additional training trajectories for the variance of the learned preference strength parameter $\beta$ under MixDPO across multiple base models and distributional parameterizations (Figure~\ref{fig:beta_trajectories}). These results complement the main text analysis in \S\ref{ssec:distributional} and illustrate how variance dynamics depend on the dataset, distributional choice and the base model.

\paragraph{Model and Parameterization Dependent Dynamics.}
Across base models (Pythia, Llama, Gemma, and Phi) and distributional parameterizations (Gamma and LogNormal), the variance of the learned preference strength exhibits model specific behavior. These differences reflect interactions between model architecture, optimization dynamics, and the chosen distributional parameterization. 

\paragraph{Dataset Level Ordering.}
Despite variation in the precise dynamics, a consistent dataset level ordering emerges across models and parameterizations. Community Alignment generally exhibits higher or more slowly decaying variance, PRISM occupies an intermediate regime, and Anthropic HH tends to concentrate variance at lower values. 

\label{sec:appendix}

\end{document}